\pdfoutput=1
\documentclass[11pt,a4paper]{article}
\usepackage[hyperref]{AACL-IJCNLP2020}
\usepackage{times}
\usepackage{latexsym}

\usepackage{CJKutf8}
\usepackage{graphics}
\usepackage{pbox}
\usepackage{makecell}
\usepackage{tabularx}
\usepackage{booktabs}

\usepackage{microtype}

\aclfinalcopy 

\title{STIL - Simultaneous Slot Filling, Translation, Intent Classification, and Language Identification: Initial Results using mBART on MultiATIS++}

\author{Jack G. M. FitzGerald \\
  Amazon Alexa AI \\
  Seattle, WA \\
  \texttt{jgmf@amazon.com}}

\date{}

\begin{document}
\maketitle
\begin{abstract}
Slot-filling, Translation, Intent classification, and Language identification, or STIL, is a newly-proposed task for multilingual Natural Language Understanding (NLU). By performing simultaneous slot filling and translation into a single output language (English in this case), some portion of downstream system components can be monolingual, reducing development and maintenance cost. Results are given using the multilingual BART model \citep{liu2020multilingual} fine-tuned on 7 languages using the MultiATIS++ dataset. When no translation is performed, mBART's performance is comparable to the current state of the art system (Cross-Lingual BERT by \citet{xu2020endtoend}) for the languages tested, with better average intent classification accuracy (96.07\% versus 95.50\%) but worse average slot F1 (89.87\% versus 90.81\%). When simultaneous translation is performed, average intent classification accuracy degrades by only 1.7\% relative and average slot F1 degrades by only 1.2\% relative.

\end{abstract}

\section{Introduction}

Multilingual Natural Language Understanding (NLU), also called cross-lingual NLU, is a technique by which an NLU-based system can scale to multiple languages. A single model is trained on more than one language, and it can accept input from more than one language during inference. In most recent high-performing systems, a model is first pre-trained using unlabeled data for all supported languages and then fine tuned for a specific task using a small set of labeled data \cite{NIPS2019_8928,pires-etal-2019-multilingual}.

\begin{table}[t]
\centering
\small
\begin{tabular}{ll}
\toprule
Input & \begin{CJK*}{UTF8}{gbsn}从 盐湖城 到 加州 奥克兰 的航班 \end{CJK*}\\ 
\midrule
\pbox{20cm}{Traditional\\Output} & \pbox{20cm}{intent: flight \\ slots:  (\begin{CJK*}{UTF8}{gbsn}盐湖城\end{CJK*}, fromloc.cityname), \\ \ldots $\;\;\;\;$(\begin{CJK*}{UTF8}{gbsn}奥克兰\end{CJK*}, toloc.cityname), \\ \ldots $\;\;\;\;$(\begin{CJK*}{UTF8}{gbsn}加州\end{CJK*}, toloc.statename) } \\
\midrule
\pbox{20cm}{STIL\\Output} & \pbox{20cm}{intent: flight \\ slots:  (salt lake city, fromloc.cityname), \\ \ldots $\;\;\;\;$(oakland, toloc.cityname), \\ \ldots $\;\;\;\;$(california, toloc.statename) \\ lang: zh} \\
\bottomrule
\end{tabular} 
\caption{\label{tab:task_example} Today's slot filling systems do not translate the slot content, as shown in ``Traditional Ouput.'' With a STIL model, the slot content is translated and language identification is performed.}
\end{table}

\begin{table*}[t]
\centering
\footnotesize
\begin{tabular}{p{0.2\textwidth}p{0.75\textwidth}}
\toprule
\textbf{Example Input} & \textbf{Example Output} \\ 
\midrule
flüge von salt lake city nach oakland kalifornien & salt $<$B-fromloc.city\_name$>$ lake $<$I-fromloc.city\_name$>$ city $<$I-fromloc.city\_name$>$ oakland $<$B-toloc.city\_name$>$ california $<$B-toloc.state\_name$>$ $<$intent-flight$>$ $<$lang-de$>$ \\
\addlinespace[0.25cm]
\begin{CJK*}{UTF8}{gbsn}从 盐湖城 到 加州 奥克兰 的航班 \end{CJK*} & salt $<$B-fromloc.city\_name$>$ lake $<$I-fromloc.city\_name$>$ city $<$I-fromloc.city\_name$>$ oakland $<$B-toloc.city\_name$>$ california $<$B-toloc.state\_name$>$ $<$intent-flight$>$ $<$lang-zh$>$ \\ 
\bottomrule
\end{tabular} 
\caption{\label{tab:inout examples} Two text-to-text STIL examples. In all STIL cases, the output is in English. Each token is followed by its BIO-tagged slot label. The sequence of tokens and slots are followed by the intent and then the language.}
\end{table*}

Two typical tasks for goal-based systems, such as virtual assistants and chatbots, are intent classification and slot filling \citep{1561278}. Though intent classification creates a language agnostic output (the intent of the user), slot filling does not. Instead, a slot-filling model outputs the labels for each of input tokens from the user. Suppose the slot-filling model can handle $L$ languages. Downstream components must therefore handle all $L$ languages for the full system to be multilingual across $L$ languages. Machine translation could be performed before the slot filling model at system runtime, though the latency would be fully additive, and some amount of information useful to the slot-filling model may be lost. Similarly, translation could occur after the slot-filling model at runtime, but slot alignment between the source and target language is a non-trivial task \citep{jain-etal-2019-entity,xu2020endtoend}. Instead, the goal of this work was to build a single model that can simultaneously translate the input, output slotted text in a single language (English), classify the intent, and classify the input language (See Table \ref{tab:task_example}). The STIL task is defined such that the input language tag is not given to the model as input. Thus, language identification is necessary so that the system can communicate back to the user in the correct language.

Contributions of this work include (1) the introduction of a new task for multilingual NLU, namely simultaneous Slot filling, Translation, Intent classification, and Language identification (STIL); (2) both non-translated and STIL results using the mBART model \citep{liu2020multilingual} trained using a fully text-to-text data format; and (3) public release of source code used in this study, with a goal toward reproducibility and future work on the STIL task\footnote{\href{https://github.com/jgmfitz/stil-mbart-multiatispp-aacl2020}{https://github.com/jgmfitz/stil-mbart-multiatispp-aacl2020}}.

\section{Dataset}

The Airline Travel Information System (ATIS) dataset is a classic benchmark for goal-oriented NLU \citep{price-1990-evaluation,Tr2010WhatIL}. It contains utterances focused on airline travel, such as \textit{how much is the cheapest flight from Boston to New York tomorrow morning?} The dataset is annotated with 17 intents, though the distribution is skewed, with 70\% of intents being the \textit{flight} intent. Slots are labeled using the Beginning Inside Outside (BIO) format. ATIS was localized to Turkish and Hindi in 2018, forming MultiATIS \citep{46604}, and then to Spanish, Portuguese, German, French, Chinese, and Japanese in 2020, forming MultiATIS++ \citep{xu2020endtoend}.

In this work, Portuguese was excluded due to a lack of Portuguese pretraining in the publicly available mBART model, and Japanese was excluded due to a current lack of alignment between Japanese and English samples in MultiATIS++. Hindi and Turkish data were taken from MultiATIS, and the training data were upsampled by 3x for Hindi and 7x for Turkish. Prior to any upsampling, there were 4,488 training samples for  English, Spanish, German, French, and Chinese. The test sets contained 893 samples for all languages except Turkish, which had 715 samples. 

For English, Spanish, German, French, and Chinese, validation sets of 490 samples were used in all cases. Given the smaller data quantities for Hindi and Turkish, two training and validation set configurations were considered. The first configuration matched that of \citet{xu2020endtoend}, using training sets of 1,495 for Hindi and 626 for Turkish along with validation sets of 160 for Hindi and 60 for Turkish. In the second configuration, no validation sets were made for Hindi and Turkish (though there were still validation sets for the other languages), and the training sets of 1,600 Hindi samples and 638 samples from MultiATIS were used.

Two output formats are considered, being (1) the non-translated, traditional case, in which translation of slot content is not performed, and (2) the translated, STIL case, in which translation of slot content is performed. In both cases, the tokens, the labels, the intent, and the detected language are all output from the model as a single ordered text sequence, as shown in Table \ref{tab:inout examples}.

\section{Related Work}

Previous approaches for intent classification and slot filling have used either (1) separate models for slot filling, including support vector machines \citep{Moschitti2007SpokenLU}, conditional random fields \citep{xu2014targeted}, and recurrent neural networks of various types \citep{Kurata_2016} or (2) joint models that diverge into separate decoders or layers for intent classification and slot filling \citep{6707709,guo2014joint,Liu_2016,hakkani-tr2016multi-domain} or that share hidden states \citep{wang-etal-2018-bi}. In this work, a fully text-to-text approach similar to that of the T5 model was used, such that the model would have maximum information sharing across the four STIL sub-tasks. 

Encoder-decoder models, first introduced in 2014 \citep{Sutskever}, are a mainstay of neural machine translation. The original transformer model included both an encoder and a decoder \citep{Vaswani}. Since then, much of the work on transformers focuses on models with only an encoder pretrained with autoencoding techniques (e.g. BERT by \citet{devlin2018bert}) or auto-regressive models with only a decoder (e.g. GPT by \citet{Radford2018ImprovingLU}). In this work, it was assumed that encoder-decoder models, such as BART \citep{lewis2019bart} and T5 \citep{raffel2019exploring}, are the best architectural candidates given the translation component of the STIL task, as well as past state of the art advancement by encoder-decoder models on ATIS, cited above. Rigorous architectural comparisons are left to future work.

\section{The Model}

\subsection{The Pretrained mBART Model}

The multilingual BART (mBART) model architecture was used \citep{liu2020multilingual}, as well as the pretrained mBART.cc25 model described in the same paper. The model consists of 12 encoder layers, 12 decoder layers, a hidden layer size of 1,024, and 16 attention heads, yielding a parameter count of 680M. The mBART.cc25 model was trained on 25 languages for 500k steps using a 1.4 TB corpus of scraped website data taken from Common Crawl \citep{wenzek2019ccnet}. The model was trained to reconstruct masked tokens and to rearrange scrambled sentences. SentencePiece tokenization \citep{Kudo_2018} was used for mBART.cc25 with a sub-word vocabulary size of 250k.

\subsection{This Work}

The same vocabulary as that of the pretrained model was used for this work, and SentencePiece tokenization was performed on the full sequence, including the slot tags, intent tags, and language tags. For all mBART experiments and datasets, data from all languages were shuffled together. The fairseq library was used for all experimentation \citep{ott-etal-2019-fairseq}.

Training was performed on 8 Nvidia V100 GPUs (16 GB) using a batch size of 32, layer normalization for both the encoder and the decoder \citep{NIPS2019_8689}; label smoothed cross entropy with $\epsilon=0.2$ \citep{Szegedy2016RethinkingTI}; the ADAM optimizer with ${\beta}_1=0.9$ and ${\beta}_2=0.999$ \citep{kingma2014adam}; an initial learning rate of $3 \times 10^{-5}$ with polynomial decay over 20,000 updates after 1 epoch of warmup; attention dropout of 0.1 and dropout of 0.2 elsewhere; and FP16 type for weights. Each model was trained for 19 epochs, which took 5-6 hours.

\begin{table*}[!t]
\centering
\small
\begin{tabular}[t]{lcccccccc}
\toprule
\textbf{Intent   accuracy}                    & \textbf{en}    & \textbf{es}    & \textbf{de}     & \textbf{zh}     & \textbf{fr}    & \textbf{hi}     & \textbf{tr}    & \textbf{Mac Avg} \\
\toprule
Cross-Lingual BERT \citep{xu2020endtoend}                   & 97.20 & 96.77 & 96.86  & 95.54  & 97.24 & \makecell[t]{92.70\\\scriptsize{tr=1495}}  & \makecell[t]{92.20\\\scriptsize{tr=626}} & 95.50     \\
Seq2Seq-Ptr \citep{10.1145/3366423.3380064} & 97.42 & & & & & & & \\
Stack Propagation \citep{qin-etal-2019-stack}                    & 97.5  &       &        &        &       &        &       &           \\
Joint BERT + CRF \citep{Chen2019BERTFJ} & 97.9 &       &        &        &       &        &       &           \\ 
\midrule
Non-translated mBART, with hi-tr val & 96.98 & 96.98 & 97.09  & 96.08  & 97.65 & \makecell[t]{95.07\\\scriptsize{tr=1495}} & \makecell[t]{92.73\\\scriptsize{tr=626}} & 96.07    \\
Translated/STIL mBART, with hi-tr val & 95.86 & 94.62 & 95.63  & 93.84  & 95.97 & \makecell[t]{93.84\\\scriptsize{tr=1495}} & \makecell[t]{91.05\\\scriptsize{tr=626}} & 94.40    \\
Non-translated mBART, no hi-tr val & 97.09 & 97.20 & 97.20  & 96.30  & 97.42 & \makecell[t]{94.74\\\scriptsize{tr=1600}} & \makecell[t]{94.27\\\scriptsize{tr=638}} & 96.32    \\
Translated/STIL mBART, no hi-tr val & 96.98 & 96.53 & 96.64  & 96.42  & 97.31 & \makecell[t]{94.85\\\scriptsize{tr=1600}} & \makecell[t]{92.87\\\scriptsize{tr=638}} & 95.94    \\
\toprule
\textbf{Slot F1}               & \textbf{en}    & \textbf{es}    & \textbf{de}     & \textbf{zh}     & \textbf{fr}    & \textbf{hi}     & \textbf{tr}    & \textbf{Mac Avg} \\
\toprule
Bi-RNN \citep{46604}                  & 95.2  &       &        &        &       & \makecell[t]{80.6\\\scriptsize{tr=600}}   & \makecell[t]{78.9\\\scriptsize{tr=600}}  & 84.90     \\
Cross-Lingual BERT \citep{xu2020endtoend}                   & 95.90 & 87.95 & 95.00  & 93.67  & 90.39 & \makecell[t]{86.73\\\scriptsize{tr=1495}} & \makecell[t]{86.04\\\scriptsize{tr=626}}  & 90.81     \\
Stack Propagation \citep{qin-etal-2019-stack}                    & 96.1  &       &        &        &       &        &       &           \\
Joint BERT \citep{Chen2019BERTFJ}                        & 96.1 &       &        &        &       &        &       &           \\
\midrule
Non-translated mBART, with hi-tr val & 95.03 & 86.76 & 94.42  & 92.13  & 89.31 & \makecell[t]{86.91\\\scriptsize{tr=1495}} & \makecell[t]{84.53\\\scriptsize{tr=626}} & 89.87    \\
Translated/STIL mBART, with hi-tr val & 93.81 & 90.38 & 91.41  & 85.93  & 91.24 & \makecell[t]{83.98\\\scriptsize{tr=1495}} & \makecell[t]{84.79\\\scriptsize{tr=626}} & 88.79    \\
Non-translated mBART, no hi-tr val & 95.00 & 86.87 & 94.14  & 92.22  & 89.32 & \makecell[t]{87.42\\\scriptsize{tr=1600}} & \makecell[t]{84.33\\\scriptsize{tr=638}} & 89.90    \\
Translated/STIL mBART, no hi-tr val & 94.66 & 91.55 & 92.61  & 87.73  & 92.15 & \makecell[t]{86.74\\\scriptsize{tr=1600}} & \makecell[t]{85.23\\\scriptsize{tr=638}} & 90.10    \\
\toprule
\textbf{Language Identification F1 }          & \textbf{en}    & \textbf{es}    & \textbf{de}     & \textbf{zh}     & \textbf{fr}    & \textbf{hi}     & \textbf{tr}    & \textbf{Mac Avg} \\
\toprule
Translated/STIL mBART, with hi-tr val & 100.00 & 98.87 & 100.00 & 100.00 & 98.95 & 100.00 & 99.93 & 99.68 \\   
Translated/STIL mBART, no hi-tr val & 99.78 & 99.83 & 100.00 & 100.00 & 99.72 & 100.00 & 99.86 & 99.88 \\   
\bottomrule
\end{tabular}
\caption{\label{tab:results} Results are shown for intent accuracy, slot F1 score, and language identification F1 score. For English, Spanish, German, Chinese, and French in all of the models shown above (including other work), training sets were between 4,478 and 4,488 samples, and validation sets were between 490 and 500 samples. In this work, two training set sizes were used for Hindi and Turkish, denoted by ``tr='' and ``with hi-tr val[idation set]'' or ``no hi-tr val[idation set]''. Across all work shown above, the tests sets contained 893 samples for all languages except Turkish, for which the test set was 715 samples.} 
\end{table*}

\section{Results and Discussion}

Results from the models are given in Table \ref{tab:results}. Statistical significance was evaluated using the Wilson method \citep{wilson} with 95\% confidence.

\subsection{Comparing to \citet{xu2020endtoend}}

Examining the first training configuration (1,496 samples for Hindi and 626 for Turkish), the non-translated mBART's macro-averaged intent classification (96.07\%) outperforms Cross-Lingual BERT by \citet{xu2020endtoend} (95.50\%), but slot F1 is worse (89.87\% for non-translated mBART and 90.81\% for Cross-Lingual BERT). The differences are statistically significant in both cases.

\subsection{With and Without Translation}

When translation is performed (the STIL task), intent classification accuracy degrades by 1.7\% relative from 96.07\% to 94.40\%, and slot F1 degrades by 1.2\% relative from 89.87\% to 88.79\%. The greatest degradation occurred for utterances involving flight number, airfare, and airport name (in that order).

\subsection{Additional Hindi and Turkish Training Data}

Adding 105 more Hindi and 12 more Turkish training examples results in improved performance for the translated, STIL mBART model. Macro-averaged intent classification improves from 94.40\% to 95.94\%, and slot F1 improves from 88.79\% to 90.10\%, both of which are statistically significant. By adding these 117 samples, the STIL mBART model matches the performance (within confidence intervals) of the non-translated mBART model. This finding suggests that the STIL models may require more training data than traditional, non-translated slot filling models.

Additionally, by adding more Hindi and Turkish data, both the intent accuracy and the slot filling F1 improves for every individual language of the translated, STIL models, suggesting that some portion of the internal, learned representation is language agnostic. 

Finally, the results suggest that there is a training-size-dependent performance advantage in using a single output language, as contrasted with the non-translated mBART model, for which the intent classification accuracy and slot F1 does not improve (with statistical significance) when using the additional Hindi and Turkish training samples.

\subsection{Language Identification}
Language identification F1 is above 99.7\% for all languages, with perfect performance in many cases. Perfect performance on Chinese and Hindi is unsurprising given their unique scripts versus the other languages tested.

\section{Conclusion}

This preliminary work demonstrates that a single NLU model can perform simultaneous slot filling, translation, intent classification, and language identification across 7 languages using MultiATIS++. Such an NLU model would negate the need for multiple-language support in some portion of downstream system components. Performance is not irreconcilably worse than traditional slot-filling models, and performance is statistically equivalent with a small amount of additional training data.
œ
Looking forward, a more challenging dataset is needed to further develop the translation component of the STIL task. The English MultiATIS++ test set only contains 455 unique entity-slot pairs. An ideal future dataset would include freeform and varied content, such as text messages, song titles, or open-domain questions. Until then, work remains to achieve parity with English-only ATIS models.

\section*{Acknowledgments}

The author would like to thank Saleh Soltan, Gokhan Tur, Saab Mansour, and Batool Haider for reviewing this work and providing valuable feedback.

{
\interlinepenalty=10000
\bibliography{bibliography}
\bibliographystyle{acl_natbib}
}

\end{document}